\documentclass[runningheads]{llncs}
\usepackage[T1]{fontenc}
\usepackage{graphicx}
\usepackage{amsmath,amssymb}
\usepackage{bm}
\usepackage{booktabs}
\usepackage{siunitx}
\usepackage{url}
\usepackage{multirow}
\usepackage{caption}

\begin{document}
\title{Has Automated Essay Scoring Reached Sufficient Accuracy? Deriving Achievable QWK Ceilings from Classical Test Theory}
\titlerunning{Has Automated Essay Scoring Reached Sufficient Accuracy?}
%
\author{Masaki Uto\orcidID{0000-0002-9330-5158}}
\authorrunning{M. Uto}
\institute{The University of Electro-Communications, Tokyo, Japan\\
\email{uto@ai.lab.uec.ac.jp}}
\maketitle              
\begin{abstract}
Automated essay scoring (AES) is commonly evaluated on public benchmarks using quadratic weighted kappa (QWK).
However, because benchmark labels are assigned by human raters and inevitably contain scoring errors, it remains unclear both what QWK is theoretically attainable and what level is practically sufficient for deployment.
We therefore derive two dataset-specific QWK ceilings based on the reliability concept in classical test theory, which can be estimated from standard two-rater benchmarks without additional annotation.
The first is the \emph{theoretical ceiling}: the maximum QWK that an ideal AES model that perfectly predicts latent true scores can achieve under label noise.
The second is the \emph{human-like ceiling}: the QWK attainable by an AES model with human-level scoring error, providing a practical target when AES is intended to replace a single human rater.
We further show that human--human QWK, often used as a ceiling reference, can underestimate the true ceiling.
Simulation experiments validate the proposed ceilings, and experiments on real benchmarks illustrate how they clarify the current performance and remaining headroom of modern AES models\footnote{This paper has been accepted at AIED 2026 as a full paper.  The final authenticated version will be published in Springer LNAI.}.
\end{abstract}

\section{Introduction}\label{sec:intro}

Automated essay scoring (AES) has attracted attention as one of the major research topics in artificial intelligence in education~\cite{ramesh2022automated,AesBook2016}.
A central goal in AES research is to improve scoring performance, and 
performance comparisons of AES models are typically conducted on public benchmark datasets using quadratic weighted kappa (QWK) as the primary evaluation metric~\cite{ijcai2019-AesSurvey,uto-review2021}.

However, even widely used benchmark datasets do not necessarily provide highly reliable score labels because labels assigned by human raters inevitably contain scoring errors due to factors such as fatigue, misunderstandings, and carelessness.
Because eliminating such errors completely is unrealistic, benchmark score labels inevitably contain a certain amount of noise.
Under this premise, even an ideal AES model that perfectly predicts the latent true score for every response would not achieve perfect agreement with the observed noisy labels (i.e., QWK would not reach 1).

Although some prior AES studies have focused on analyzing label noise in datasets or mitigating its effects to improve robustness~\cite{Amorim2018AutomatedES,uto2022learning,Wind2018}, prior work has not sufficiently discussed attainable QWK ceilings that explicitly account for label reliability.
Without such dataset-specific ceilings, the remaining headroom becomes ambiguous, which may lead to an overemphasis on trivial differences among models.
Moreover, for practical deployment, it is often more important to interpret whether an AES model has reached a human-replaceable level of agreement than to compare marginal differences in QWK.

To address these issues, we derive two dataset-specific QWK ceilings based on the reliability concept in classical test theory (CTT), under a standard setting in which each response has scores from two human raters.
Specifically, we propose the following two ceilings.
\begin{enumerate}
\item {\bf Theoretical ceiling:} the maximum QWK that cannot be exceeded because of the reliability limitation of the score labels, even by an ideal AES model that perfectly predicts each response's true score. This ceiling provides a principled upper bound on attainable QWK.
\item {\bf Human-like ceiling:} the QWK attainable by an AES model whose scoring error is comparable to that of a single human rater. This ceiling provides a practical target when AES is intended to replace one human rater.
\end{enumerate}

Furthermore, we show that human--human QWK, which has often been used as a ceiling reference in prior studies~\cite{chen-he-2013-automated,cozma-etal-2018-automated,taghipour-ng-2016-neural}, can be a conservative reference that underestimates the true ceiling.

The main contributions of this study are as follows.
\begin{enumerate}
\item By quantifying dataset-specific QWK ceilings, we enable the interpretation of AES performance not only through relative comparisons but also as an absolute degree of attainment, thereby clarifying the current status and supporting deployment decisions.
\item By clarifying that human--human QWK can underestimate the true ceiling, we correct a potentially misleading ceiling reference.
\item By evaluating modern AES models on representative benchmarks using the proposed ceilings as a common yardstick, we provide an interpretation of their current status and remaining headroom.
\end{enumerate}

\section{Automated Essay Scoring Technologies}\label{sec:aes}

Automated scoring of constructed responses has evolved into two related areas: AES and automated short answer grading (ASAG)~\cite{Burrows2015,ijcai2019-AesSurvey,AesBook2016}.
This study focuses primarily on AES for simplicity, although our framework is also applicable to ASAG in the same manner.

Early AES models relied on manually engineered features combined with regression or classification models~\cite{AesBook2016}.
While that approach offered interpretability, achieving high performance typically required dataset-specific feature engineering.
In contrast, deep neural network (DNN)-based methods that automatically learn feature representations have developed rapidly since the 2010s~\cite{elmassry2025systematic,ijcai2019-AesSurvey,misgna2024survey,ramesh2022automated,uto-review2021}.
In particular, fine-tuning pretrained Transformer encoders such as BERT (Bidirectional Encoder Representations from Transformers) on scored essay data has become a standard baseline for DNN-based AES~\cite{elmassry2025systematic,misgna2024survey,ramesh2022automated,uto-review2021,yamaura2023neural}.
More recently, researchers have actively studied a new paradigm in which large language models (LLMs) are used to perform scoring with either no or only a small amount of training data~\cite{lee2024unleashing,mansour2024can,yancey2023rating}.
While LLM-based AES is promising when training data are scarce, studies have reported that when sufficient training data are available, fine-tuned BERT-based and related AES models still tend to achieve higher performance~\cite{lee2024unleashing,mansour2024can,shibata2025cross,yancey2023rating}.

AES performance is typically evaluated on public benchmark datasets using hold-out evaluation or cross-validation, with QWK as the standard agreement metric~\cite{ijcai2019-AesSurvey,misgna2024survey,uto-review2021}.
However, prior work has mainly focused on relative comparisons across AES methods, and discussion of absolute attainment targets has been limited.
While human--human QWK has often been used as a reference value~\cite{chen-he-2013-automated,cozma-etal-2018-automated,taghipour-ng-2016-neural}, its validity has not been thoroughly discussed.
Accordingly, the aims of the present study are to define dataset-specific attainable QWK ceilings and provide an interpretation of human--human QWK.

\section{Task Setting} \label{sec:assumption}

In AES benchmark datasets, the target variable for each response is often defined as the mean or sum of scores assigned by multiple human raters (annotators).
In this study, we assume this setting and derive the proposed ceilings based on the reliability concept in CTT.

Following standard CTT assumptions, we model the score $X_{ij}$ assigned by rater $j$ to response $i$ as the sum of a latent true score and random error as follows:
\begin{equation}
X_{ij} = T_i + \varepsilon_{ij}.
\label{eq:ctt_model}
\end{equation}
Here, $T_i$ denotes the latent true score of response $i$, which can be regarded as the expectation of scores over a rater population, and $\varepsilon_{ij}$ denotes an error term.
As standard CTT assumptions on errors, we assume $\mathbb{E}[\varepsilon_{ij}] = 0$, $\mathrm{Cov}(T_i, \varepsilon_{ij}) = 0$, and $\mathrm{Cov}(\varepsilon_{ij}, \varepsilon_{ij'}) = 0$ for $j \neq j'$.
Here, $\mathbb{E}[\cdot]$ denotes expectation, and $\mathrm{Cov}(\cdot,\cdot)$ denotes covariance between two random variables.

We further assume that the target variable is the mean of the observed scores from $J$ raters:
\begin{equation}
Y_{i} = \frac{1}{J} \sum_{j=1}^{J} X_{ij}.
\label{eq:mean_score}
\end{equation}
Note that although the following discussion proceeds under the mean-score setting, it also applies directly to the sum of the $J$ rater scores because the sum is a constant multiple of the mean and our analysis is invariant to linear transformations.
We further focus on the case of $J=2$ because it is both analytically convenient and common in AES benchmarks, although our framework can be generalized to $J>2$.

Next, we assume that the goal of AES is to predict each response's true score $T$ by learning from training data whose target variable is $Y$.
This can be justified from the perspective of statistical learning.
Standard AES models are often trained by minimizing mean squared error (MSE), which is equivalent to estimating the conditional expectation $\hat{Y} = \mathbb{E}[Y \mid \bm{Z}]$ given the input essay $\bm{Z}$~\cite{Bishop:2006}.
Under the above CTT assumptions, the expected value of the observed target score equals the true score, i.e., $\mathbb{E}[Y \mid T] = T$.
Therefore, learning a regression model for $Y$ can be viewed as learning a predictor of the latent true score $T$.

Now assume that we can construct an ideal AES model that predicts $T$ without error, and that its performance is evaluated by comparing its predictions with the target variable $Y$ on an evaluation dataset.
When QWK is used as the evaluation metric, the attainable QWK is constrained by the magnitude of the error component inherent in $Y$.
The key idea of this study is to estimate this error component via the reliability concept in CTT and then derive the attainable QWK under that noise.

\section{Proposed Ceilings}

In this section, we derive the two ceilings following the procedures described above.
We derive the theoretical ceiling in Section~\ref{sec:qwk_ub} and the human-like ceiling in Section~\ref{sec:human_like_ceiling}, then in Section~\ref{sec:human_qwk} we clarify the relationships among the proposed ceilings and human--human QWK.

\subsection{Deriving the Theoretical Ceiling}\label{sec:qwk_ub}

As discussed above, the theoretical ceiling of QWK depends on the magnitude of the error component in the target variable $Y$.
In CTT, the proportion of error variance contained in an observed score is quantified by the concept of reliability.

From Eqs.~\eqref{eq:ctt_model} and \eqref{eq:mean_score}, the target variable $Y$ can be decomposed as
\begin{equation}
Y_i  = \frac{1}{J}\sum_{j=1}^J X_{ij} = T_i + \bar{\varepsilon}_i, \quad \bar{\varepsilon}_i = \frac{1}{J}\sum_{j=1}^J \varepsilon_{ij}.
\label{eq:Y_decomp}
\end{equation}
Under this decomposition, the reliability of $Y$ in CTT is defined as
\begin{equation}
\rho_{Y} = \frac{\sigma^2_{T}}{\sigma^2_{Y}} = \frac{\sigma^2_{T}}{\sigma^2_{T} + \sigma^2_{\bar{\varepsilon}}}, \label{eq:reliability_def_Y}
\end{equation}
where $\sigma^2_{Y}$, $\sigma^2_{T}$, and $\sigma^2_{\bar{\varepsilon}}$ denote the variances of $Y$, $T$, and $\bar{\varepsilon}$.
This definition implies that reliability approaches $1$ as the error variance $\sigma^2_{\bar{\varepsilon}}$ approaches $0$.

Because the true-score variance $\sigma^2_T$ is not directly observable, we cannot compute $\rho_Y$ exactly from Eq.~\eqref{eq:reliability_def_Y}.
Therefore, we estimate reliability from observable data.
When raters are randomly assigned and rater identities are unavailable for each response, as in many public benchmark datasets, the reliability estimate $\hat{\rho}_Y$ can be obtained as an intraclass correlation coefficient under a one-way random-effects, average-measure model~\cite{mcgraw1996forming}.

Using the reliability estimate $\hat{\rho}_Y$, we derive the theoretical ceiling by considering an ideal AES model whose output predictions are perfectly correlated with the true score (i.e., $\hat{Y}\propto T$) and that is calibrated to match the mean and variance of the target variable (i.e., $\mu_{\hat{Y}}=\mu_Y$ and $\sigma_{\hat{Y}}=\sigma_Y$).
Because direct derivation of the QWK ceiling is inconvenient, we first derive the ceiling of the correlation coefficient.

Under the ideal AES model, the maximum attainable correlation $r_{\max}$ between the model prediction $\hat{Y}$ and the target variable $Y$ is given as follows (see Appendix~A for the proof):
\begin{equation}
r_{\max} = \sqrt{\rho_{Y}}.
\label{eq:rmax}
\end{equation}
This equation indicates that even for the ideal AES model, the observed correlation with the noisy target variable cannot reach $1$ and is capped at $\sqrt{\rho_Y}$.
Accordingly, $\sqrt{\rho_Y}$ can be interpreted as the theoretical ceiling of the correlation coefficient, and we can estimate it as $\hat{r}_{\max}=\sqrt{\hat{\rho}_{Y}}$.

We then derive the QWK ceiling by converting this correlation ceiling into QWK.
For this purpose, we use the following QWK approximation based on Lin's concordance correlation coefficient (CCC)~\cite{barnhart2007overall}:
\begin{equation}
\kappa \approx r \left( \frac{2 \sigma_{\hat{Y}} \sigma_{Y}}{\sigma_{\hat{Y}}^2 + \sigma_{Y}^2 + (\mu_{\hat{Y}} - \mu_{Y})^2} \right),
\label{eq:qwk_max}
\end{equation}
where $\kappa$ indicates the approximated QWK value, $r$ denotes the correlation between $\hat{Y}$ and $Y$, $\mu_{\hat{Y}}$ and $\sigma^2_{\hat{Y}}$ are respectively the mean and variance of $\hat{Y}$, and $\mu_Y$ and $\sigma^2_Y$ are respectively the mean and variance of $Y$.

Under the ideal AES model that attains $r_{\max}$ and satisfies $\mu_{\hat{Y}}=\mu_Y$ and $\sigma_{\hat{Y}}=\sigma_Y$, the CCC-based approximation implies that the corresponding QWK can be computed as
\begin{equation}
\kappa_{\max} \approx r_{\max} = \sqrt{\rho_{Y}}.
\end{equation}
Accordingly, the theoretical ceiling of QWK can be estimated as $\hat{\kappa}_{\max}=\sqrt{\hat{\rho}_Y}$.

Note that the CCC-based approximation in Eq.~\eqref{eq:qwk_max} provides a close approximation to QWK under quadratic weights on an equally spaced score scale~\cite{barnhart2007overall}.
Meanwhile, in AES studies, scores are typically discrete and bounded, and the target score is often computed as the rounded average of multiple rater scores, which can introduce quantization noise.
Therefore, in Section~\ref{sec:simulation} we confirm empirically that the CCC-based ceiling remains accurate under such AES settings.

\subsection{Deriving the Human-like Ceiling}\label{sec:human_like_ceiling}

In this subsection, we derive the second ceiling value, the human-like ceiling, which is relevant when AES is intended to substitute for a single human rater.
In practical deployment, AES is often introduced not to fully automate scoring but rather to serve as one side of double scoring or to verify human scoring~\cite{EnrightQuinlan2010,RamineniWilliamson2013}.
In such situations, a more direct and reasonable reference is the QWK attainable by an AES model whose scoring error is comparable to that of a single human rater (hereafter referred to as a \emph{human-like model}).

Accordingly, we define the human-like ceiling as the QWK that a human-like model can achieve against the target variable $Y$.
The derivation follows the same idea as for the theoretical ceiling. 
We first derive the correlation implied by single-rater reliability and then convert it into a QWK ceiling via the CCC approximation.

Let the output of a human-like model be $\tilde{X}$, and assume that $\tilde{X}$ measures the true score $T$ while having an error variance comparable to that of a single-rater score $X$:
\begin{equation}
\tilde{X}_{i} = T_i + \eta_{i},
\end{equation}
where $\mathbb{E}[\eta_{i}] = 0$, $\mathrm{Cov}(T_i, \eta_{i}) = 0$, $\mathrm{Cov}(\varepsilon_{ij}, \eta_{i}) = 0$, and $\sigma^2_{\eta} = \sigma^2_{\varepsilon}$.
Here, $\sigma^2_{\eta}$ denotes the variance of $\eta_i$.

Under this assumption, the reliability of $\tilde{X}$ is equivalent to the reliability of a single-rater observed score $X$, given by
\begin{equation}
\rho_{1} = \frac{\sigma_T^2}{\sigma_X^2} = \frac{\sigma_T^2}{\sigma_T^2+\sigma_{\varepsilon}^2},\label{eq:reliability_def_1}
\end{equation}
where $\sigma^2_{\varepsilon}$ denotes the variance of the error $\varepsilon$.
In this setup, an estimator $\hat{\rho}_1$ can be obtained as an intraclass correlation coefficient under a one-way random-effects single-measure model~\cite{mcgraw1996forming}.

As shown in Appendix~A, the correlation between $\tilde{X}$ and the target variable $Y$ is derived as follows:
\begin{equation}
r_{\mathrm{HL}} = \sqrt{\rho_{1}\rho_{Y}}.
\label{eq:cor_humanlike}
\end{equation}
Therefore, when the mean and variance of $\tilde{X}$ are aligned with those of $Y$, the CCC approximation yields
\begin{equation}
\kappa_{\mathrm{HL}} \approx r_{\mathrm{HL}} = \sqrt{\rho_{1}\rho_{Y}}.
\label{eq:kappa_humanlike}
\end{equation}
Accordingly, $\kappa_{\mathrm{HL}}$ can be interpreted as the human-like ceiling, and it can be estimated as $\hat{\kappa}_{\mathrm{HL}}=\sqrt{\hat{\rho}_1\hat{\rho}_Y}$.

\subsection{Interpreting Human--Human QWK}\label{sec:human_qwk}

As mentioned in Sections~\ref{sec:intro} and~\ref{sec:aes}, prior studies have often used human--human QWK (denoted by $\kappa_{H}$) as an empirical reference ceiling.
Here, we show that this can be a conservative reference that underestimates the true ceiling.

Let $r_H$ denote the correlation between two sequences of human scores $(X_{\cdot 1},X_{\cdot 2})$.
As shown in Appendix~A, the following inequality holds:
\begin{equation}
    r_H \le r_{\max}.
    \label{eq:rH_le_rmax}
\end{equation}
Furthermore, using the CCC-based approximation of QWK, we obtain
\begin{equation}
    \kappa_H \approx r_H\,F_H \le r_H.
    \label{eq:kappaH_le_rH}
\end{equation}
Here, $F_H$ is defined as
\begin{equation}
F_H = \frac{2 \sigma_{1} \sigma_{2}}{\sigma_{1}^2 + \sigma_{2}^2 + (\mu_{1} - \mu_{2})^2},
\end{equation}
where $\mu_{j}$ and $\sigma_{j}^2$ denote the mean and variance of the $j$-th human score sequence.
Note that the inequality $r_H\,F_H \le r_H$ in Eq.~\eqref{eq:kappaH_le_rH} follows because $\sigma_{1}^2 + \sigma_{2}^2 \ge 2 \sigma_{1} \sigma_{2}$ implies $0 \le F_H \le 1$, and typically $r_H \ge 0$.

From these results and the relation $r_{\max} \approx \kappa_{\max}$, we obtain
\begin{equation}
    \kappa_H \lesssim r_H \le r_{\max} \approx \kappa_{\max}.
    \label{eq:kappaH_pessimistic}
\end{equation}
This indicates that human--human QWK is generally lower than the theoretical ceiling derived in this study.
This is intuitive because agreement between an AES prediction trained to predict an average of two raters and the averaged target score, which contains less noise, is typically higher than the agreement between two single-rater scores that each contain more noise.

Moreover, the human-like ceiling $\kappa_{\mathrm{HL}}$ provides a criterion between human--human QWK $\kappa_H$ and the theoretical ceiling $\kappa_{\max}$.
Specifically, as shown in Appendix~A, the following relationship typically holds:
\begin{equation}
    \kappa_H \lesssim \kappa_{\mathrm{HL}}\le \kappa_{\max}. \label{eq:all_index}
\end{equation}
Accordingly, $\kappa_{\mathrm{HL}}$ can serve as an operational target in cases where an AES surpasses human--human agreement but does not reach the theoretical ceiling.

Furthermore, we can expect the gap among the three ceiling/reference values to increase as rater noise increases.
This trend can be explained by comparing the definitions of $\rho_Y$ and $\rho_1$.
From Eqs.~\eqref{eq:reliability_def_Y} and~\eqref{eq:reliability_def_1}, the key difference is the error variance in the denominator:
$\rho_1$ involves $\sigma^2_{\varepsilon}$, whereas $\rho_Y$ involves $\sigma^2_{\bar{\varepsilon}}$.
Moreover, when errors are independent and homoscedastic, we have $\sigma^2_{\bar{\varepsilon}}=\sigma^2_{\varepsilon}/J$ (i.e., $\sigma^2_{\varepsilon}/2$ for two raters).
These observations imply that $\rho_1$ decreases more rapidly than does $\rho_Y$ as $\sigma^2_{\varepsilon}$ increases.
Therefore, the three values respond differently to increasing rater noise.
$\hat{\kappa}_{\max} \approx \sqrt{\hat{\rho}_Y}$ is the most robust, $\hat{\kappa}_{\mathrm{HL}} \approx \sqrt{\hat{\rho}_1\hat{\rho}_Y}$ is intermediate, and $\hat{\kappa}_{H}$ decreases the most because $\hat{\kappa}_{H} \le \hat{\rho}_1$ as shown in Appendix~A.
These results suggest that in datasets with larger rater scoring errors, human--human QWK can underestimate the true ceiling by a larger margin.

\section{Validation via Simulation Experiments}
\label{sec:simulation}

We conducted simulation experiments to validate the proposed ceilings.
In the simulations, we generated latent true scores, then generated observed scores by adding controlled errors, and used the resulting data to verify the following three points.
\begin{enumerate}
    \item Whether the empirical QWK between the latent true scores and the observed mean scores agrees well with the proposed theoretical ceiling.
    \item Whether the results are consistent with the theoretical relationships among the ceilings in Section~\ref{sec:human_qwk} (i.e., $\kappa_H \lesssim \kappa_{\mathrm{HL}}\le \kappa_{\max}$).
    \item Whether the CCC-based approximation of QWK remains sufficiently accurate under AES-like settings with discrete, bounded score scales and rounding-induced quantization.
\end{enumerate}

\subsection{Validating the Proposed Ceilings}\label{sec:simulation_procedure}

In this experiment, we set the sample size (the number of responses) to $N=1000$ and the score range to integers from $0$ to $10$, and generated true scores $T_i$.
Specifically, we sampled values from a normal distribution with mean $5.0$ and standard deviation $3.3$, rounded them to the nearest integer, and clipped them to $[0,10]$:
\begin{equation}
    T_{i} = \text{clip}(\text{round}(T^{\prime}_i)), \quad T^{\prime}_i \sim N(5.0, 3.3^2).
\end{equation}
Here, $N(\mu, \sigma^2)$ denotes a normal distribution with mean $\mu$ and variance $\sigma^2$, $\text{round}(\cdot)$ denotes rounding to the nearest integer, and $\text{clip}(\cdot)$ denotes clipping to the interval $[0, 10]$.

Next, we generated observed scores $X_{i1}$ and $X_{i2}$ from two raters by adding independent and identically distributed errors to the true scores:
\begin{equation}
    X_{ij} = \text{clip}(\text{round}(T_i + \varepsilon_{ij})), \quad \varepsilon_{ij} \sim N(0, \sigma_{noise}^2),
\end{equation}
where $\sigma_{noise}^2$ denotes the error variance.
We then constructed the target variable as the rounded mean of the two raters' scores:
$Y_i = \mathrm{round}((X_{i1} + X_{i2})/2)$.

We repeated the above data generation 100 times for each of five noise levels $\sigma_{noise} \in \{0.25, 0.50, 1.00, 2.00, 3.00\}$.
For each trial, we computed (i) the correlation and QWK between the ideal AES prediction $\hat{Y}_i = T_i$ and the target variable $Y$ (denoted by $r_{\mathrm{true}}$ and $\kappa_{\mathrm{true}}$), (ii) the theoretical ceiling ($\hat{\kappa}_{\max}$), (iii) the human-like ceiling ($\hat{\kappa}_{\mathrm{HL}}$), and (iv) the human--human QWK ($\hat{\kappa}_{H}$).
For each value, we reported the mean over 100 trials.

\begin{table*}[t]
  \centering
  \begin{minipage}[c]{0.48\textwidth}
    \centering
    \caption{Simulation results.}
    \label{tab:simulation_ideal}
    \setlength{\tabcolsep}{3pt} 
    \begin{tabular}{cccccc} %
      \toprule
      $\sigma_{noise}$ & $r_{\mathrm{true}}$ & $\kappa_{\mathrm{true}}$ & $\hat{\kappa}_{\max}$ & $\hat{\kappa}_{\mathrm{HL}}$ & $\hat{\kappa}_{H}$ \\
      \midrule
      0.25 & 0.993 & 0.990 & 0.996 & 0.989 & 0.985\\
      0.50 & 0.987 & 0.982 & 0.991 & 0.974 & 0.965\\
      1.00 & 0.967 & 0.961 & 0.972 & 0.919 & 0.895\\
      2.00 & 0.897 & 0.890 & 0.901 & 0.749 & 0.683\\
      3.00 & 0.802 & 0.797 & 0.806 & 0.563 & 0.482\\
      \bottomrule
    \end{tabular}
  \end{minipage}
  \hfill %
  \begin{minipage}[c]{0.48\textwidth}
    \centering
    \includegraphics[width=.65\linewidth]{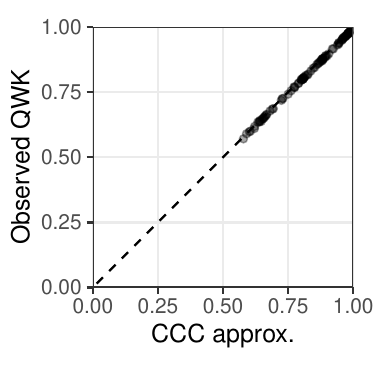}
    \captionof{figure}{Plot of QWK and CCC.}
    \label{fig:sim_qwk_vs_ccc}
  \end{minipage}
\end{table*}

The results are given in Table~\ref{tab:simulation_ideal}, where the values of $\kappa_{\mathrm{true}}$ closely match the estimated theoretical ceiling $\hat{\kappa}_{\max}$ across all noise levels, and $r_{\mathrm{true}}$ exhibits a similar pattern.
This indicates that the proposed theoretical ceiling provides an accurate estimate of the attainable QWK.

Next, comparing the three ceiling/reference values, we observe that $\hat{\kappa}_H \le \hat{\kappa}_{\mathrm{HL}}\le \hat{\kappa}_{\max}$ holds in all conditions.
This supports our theoretical claim that human--human QWK is not a logical ceiling on AES performance, but rather a conservative reference based on agreement between noisy observations.

We also observe that the gap among the three values tends to increase as the noise level increases.
This trend is consistent with the theoretical discussion in Section~\ref{sec:human_qwk}, further supporting the validity of the proposed ceilings.

\subsection{Accuracy of the CCC-Based QWK Approximation}

We conducted another simulation experiment to evaluate the accuracy of the CCC-based QWK approximation.
Specifically, we repeated the simulation procedure in Section~\ref{sec:simulation_procedure} 100 times while randomly sampling the rater noise level $\sigma_{noise}$ from a uniform distribution $U(0.1, 5)$.
For each trial, we computed the CCC-approximated QWK, denoted by $\kappa_{\mathrm{ccc}}$, using the correlation between $T$ and $Y$ and the means and variances of the two score variables, along with the true QWK, denoted by $\kappa_{\mathrm{true}}$, computed directly from $T$ and $Y$.
We then computed the approximation error as $e=\kappa_{\mathrm{true}}-\kappa_{\mathrm{ccc}}$.

Figure~\ref{fig:sim_qwk_vs_ccc} shows $\kappa_{\mathrm{ccc}}$ (horizontal axis) plotted against $\kappa_{\mathrm{true}}$ (vertical axis), showing that the points lie close to the line of equality (dashed).
Furthermore, the mean absolute error $\mathbb{E}[|e|]$ over the 100 trials was 0.005.
These results indicate that the CCC-based approximation of QWK is sufficiently accurate for our AES settings.

\section{Validation on Real Data}
\label{sec:real_data}

In this section, we compute the proposed ceilings on the following two benchmark datasets and compare them with the QWK of modern AES models.

\begin{description}
    \item[ASAP (Automated Student Assessment Prize).] 
    This dataset consists of scored English essays for eight prompts; each prompt contains between 723 and 1805 essays.
    The score range varies across prompts, from 4 to 61 score levels.
    Each essay has scores from two raters.
    Following common practice and our assumptions in Section~\ref{sec:assumption}, we constructed the target variable as the mean of the two raters' scores for Prompts~2--6, and as the simple sum for Prompts~1, 7, and~8.
    \item[ELLIPSE~\cite{ellipse_aied24}.] 
    This dataset consists of essays written by English learners, with scores assigned separately for six traits (e.g., Cohesion and Syntax).
    The dataset provides two raters' scores per essay and an official split into 3911 training essays and 2571 test essays.
    We constructed the target variable for each trait as the mean of the two raters' scores.
\end{description}
Note that several details regarding the above target construction and the rater assignment design are discussed in Appendix~B.

For each dataset, we computed the theoretical ceiling $\kappa_{\max}$, the human-like ceiling $\kappa_{\mathrm{HL}}$, and the human--human QWK $\kappa_{H}$.
Specifically, we computed these values using all essays for each prompt in ASAP, and using the official test split for each trait in ELLIPSE.

We also evaluated a BERT-based AES model with a standard architecture~\cite{uto-review2021} to examine how close modern AES models are to the proposed ceilings.
For ASAP, we performed five-fold cross-validation for each prompt, concatenated the out-of-fold predictions, and computed correlation and QWK against the corresponding target scores.
For ELLIPSE, we trained the model on the official training split for each trait and computed correlation and QWK on the official test split.
For QWK, we rounded and clipped both model predictions and target scores to the valid score range. 

Furthermore, as stronger baselines than BERT, we also report results from NPCR (Neural Pairwise Contrastive Regression)~\cite{npcr2022} for ASAP and MTAA (Multi-Task Automated Assessment)~\cite{ellipse_aied24} for ELLIPSE.
Note that these values are not strictly comparable to our BERT baseline because of differences in experimental setups; see Appendix~B for details.

\begin{table*}[t]
\setlength{\tabcolsep}{5pt} 
\centering
\caption{Analysis results on the ASAP dataset.}
\label{tab:asap_results}
\begin{tabular}{lccccccc}
\toprule
\multirow{2}{*}{Prompt} & \multirow{2}{*}{$\hat{\kappa}_{\max}$} & \multirow{2}{*}{$\hat{\kappa}_{\mathrm{HL}}$} & \multirow{2}{*}{$\hat{\kappa}_{H}$} & \multicolumn{2}{c}{BERT} & NPCR \\ \cline{5-6}
  &  &  & & QWK & Correl & (QWK) \\
\midrule
Prompt 1 & 0.915 & 0.777 & 0.721 & 0.803 & 0.809 & 0.856 \\
Prompt 2 & 0.947 & 0.855 & 0.814 & 0.698 & 0.704 & 0.750 \\
Prompt 3 & 0.932 & 0.818 & 0.769 & 0.647 & 0.648 & 0.756 \\
Prompt 4 & 0.959 & 0.885 & 0.851 & 0.764 & 0.764 &  0.851 \\ 
Prompt 5 & 0.927 & 0.804 & 0.753 & 0.779 & 0.781 &  0.847 \\ 
Prompt 6 & 0.935 & 0.824 & 0.776 & 0.777 & 0.787 &  0.858 \\ 
Prompt 7 & 0.916 & 0.778 & 0.721 & 0.824 & 0.826 &  0.838 \\ 
Prompt 8 & 0.879 & 0.697 & 0.629 & 0.723 & 0.749 &  0.779 \\ 
\hline
Avg. & 0.926  & 0.805 & 0.754 & 0.752 & 0.759 & 0.817 \\
\bottomrule
\end{tabular}
\end{table*}

\begin{table*}[t]
\setlength{\tabcolsep}{5pt} 
\centering
\caption{Analysis results on the ELLIPSE dataset.}
\label{tab:ellipse_results}
\begin{tabular}{lcccccccccc}
\toprule
\multirow{2}{*}{Trait} & \multirow{2}{*}{$\hat{\kappa}_{\max}$} & \multirow{2}{*}{$\hat{\kappa}_{\mathrm{HL}}$} & \multirow{2}{*}{$\hat{\kappa}_{H}$} & & \multicolumn{2}{c}{BERT} & & \multicolumn{2}{c}{MTAA} \\ \cline{6-7}\cline{9-10}
 &  &  & & & QWK & Correl & & QWK & Correl \\
\midrule
Cohesion & 0.818 & 0.581 & 0.504 & & 0.476 & 0.505 & & 0.630 & 0.660 \\
Syntax & 0.817 & 0.578 & 0.501 & & 0.533 & 0.575 & & 0.690 & 0.720 \\
Vocabulary & 0.799 & 0.547 & 0.469 & & 0.534 & 0.557 & & 0.670 & 0.700 \\
Phraseology & 0.814 & 0.573 & 0.496 & & 0.531 & 0.586 & & 0.690 & 0.730 \\
Grammar & 0.828 & 0.597 & 0.522 & & 0.592 & 0.613 & & 0.720 & 0.740 \\
Conventions & 0.823 & 0.588 & 0.511 & & 0.564 & 0.591 & & 0.710 & 0.750 \\
\hline
Avg. & 0.816 & 0.577 & 0.501 & & 0.538 & 0.571 & & 0.685 & 0.717 \\
\bottomrule
\end{tabular}
\end{table*}

The results are given in Tables~\ref{tab:asap_results} and~\ref{tab:ellipse_results}.
Because NPCR did not report correlation coefficients, we report only QWK for NPCR.
First, comparing the three values, we observe that $\hat{\kappa}_H \le \hat{\kappa}_{\mathrm{HL}}\le \hat{\kappa}_{\max}$ holds in all conditions, consistent with the theory.
Moreover, for ELLIPSE where human--human QWK is relatively low, the gap between $\hat{\kappa}_{\max}$ and $\hat{\kappa}_H$ is large, suggesting that the conventional reference $\hat{\kappa}_H$ can substantially underestimate the true ceiling.
Furthermore, the BERT and MTAA results confirm that QWK is generally lower than correlation, as expected from Eq.~\eqref{eq:kappaH_le_rH}.
Overall, these findings indicate that our theoretical expectations align well with real-data observations.

Next, when comparing AES performance with the proposed ceilings, we observe multiple cases where model performance exceeds $\hat{\kappa}_{H}$.
For example, for ASAP Prompt~1, $\hat{\kappa}_{H}$ is 0.721, whereas BERT achieves 0.803 and NPCR achieves 0.856, substantially exceeding human--human agreement.
For ELLIPSE, the average performance of both models exceeds $\hat{\kappa}_{H}$.
These results confirm that human--human QWK is a conservative reference that can be surpassed in practice.

Meanwhile, across all conditions, current models still show a nontrivial gap to the theoretical ceiling $\hat{\kappa}_{\max}$.
In contrast, the human-like ceiling $\hat{\kappa}_{\mathrm{HL}}$ is achieved or exceeded by BERT in some conditions, and by NPCR and MTAA in many cases.
These observations suggest that while substantial headroom remains with respect to the theoretical ceiling, current AES models may already reach a level of agreement comparable to that of a single human rater in many settings.

\section{Conclusion}\label{sec:conclusion}

In this study, we presented a framework for quantifying dataset-specific QWK ceilings by leveraging noise in score labels estimated via the reliability concept in CTT.
As one ceiling, we derived the theoretical ceiling ($\kappa_{\max}$), defined as the maximum QWK that even an ideal AES model can attain under the reliability constraint of the benchmark labels.
As another ceiling, we derived the human-like ceiling ($\kappa_{\mathrm{HL}}$), defined as the QWK attainable by an AES model with scoring error comparable to that of a single human rater.
We also theoretically positioned human--human QWK ($\kappa_H$) as a conservative reference rather than a true ceiling.

Via simulations and real-data experiments, we showed that (i) the proposed theoretical ceiling estimate closely matches the true QWK ceiling, (ii) the relationship $\kappa_H \lesssim  \kappa_{\mathrm{HL}} \le \kappa_{\max}$ holds as expected, (iii) the CCC-based approximation of QWK yields sufficiently small error, and (iv) on two benchmark datasets, modern AES models often exceed $\kappa_H$ while remaining below $\kappa_{\max}$, and in some cases reach or surpass $\kappa_{\mathrm{HL}}$.
Overall, these results suggest that the proposed ceilings provide a coherent interpretation and an absolute yardstick for evaluating AES performance on each dataset.

While this study provides a new perspective for AES research, it focused on datasets where the target variable is constructed as the mean or sum of multiple rater scores, and it relied on standard CTT assumptions such as independence and homogeneity of rater errors.
However, violations of these assumptions can affect the interpretation of the proposed QWK ceilings. For example, if raters share biases in the same direction, the actual ceiling attainable by an ideal AES model may be smaller than the proposed ceilings. This is because such shared error is not sufficiently reduced by averaging across raters, so the actual target scores can contain more residual noise than is assumed in the current framework. Because our theory can in principle be extended by relaxing these assumptions, an important direction for future work is to generalize the framework so that it can be applied to a broader variety of datasets and scoring settings. In this direction, extensions based on more general psychometric frameworks, such as generalizability theory~\cite{brennan2001generalizability} and many-facet Rasch models~\cite{MFRM2016BOOK,uto2020item}, may also be considered, although such approaches typically require additional information about the scoring design, such as rater identities and how raters are assigned to responses, which is not always available in standard AES benchmarks. In addition, quantifying uncertainty in the estimated ceiling values, for example through bootstrap-based confidence intervals, is also an important topic for future work. Another important topic for future work is to examine the accuracy of the CCC-based QWK approximation under more extreme boundary conditions, such as highly skewed score distributions, strong floor or ceiling effects, or very coarse score scales.

\appendix

\section{Derivations and Proofs}

Equation~\eqref{eq:rmax} can be proved under the assumptions in Section~\ref{sec:assumption} as follows:
\begin{equation}
r_{\max} = \mathrm{Cor}(\hat{Y}, Y) = \mathrm{Cor}(T, Y) 
 = \frac{\mathrm{Cov}(T,T+\bar{\varepsilon})}{\sqrt{{\sigma^2_T}{\sigma^2_Y}}} 
 = \frac{\sigma^2_T}{\sqrt{{\sigma^2_T}{\sigma^2_Y}}} = \sqrt{\rho_{Y}},
\end{equation}
where $\mathrm{Cor}(T, Y)$ denotes the correlation between $T$ and $Y$.

Furthermore, Eq.~\eqref{eq:cor_humanlike} in Section~\ref{sec:human_like_ceiling} can be derived under the assumptions in Section~\ref{sec:assumption} as follows:
\begin{equation}
r_{\mathrm{HL}}=\mathrm{Cor}(\tilde{X},Y)
= \frac{\mathrm{Cov}(T+\eta,\,T+\bar{\varepsilon})}
        {\sqrt{\mathrm{Var}(T+\eta)\,\mathrm{Var}(T+\bar{\varepsilon})}}
= \frac{\sigma_T^2}
        {\sqrt{(\sigma_T^2+\sigma^2_{\eta})(\sigma_T^2+\sigma^2_{\bar{\varepsilon}})}} = \sqrt{\rho_1\rho_Y}.
\end{equation}
Here, $\mathrm{Var}(\cdot)$ denotes variance.

Next, we prove Eq.~\eqref{eq:rH_le_rmax} in Section~\ref{sec:human_qwk} for the case $J=2$.
To show this, it suffices to show $\rho_Y - r_H^2 \ge 0$.
Under the assumptions in Section~\ref{sec:assumption}, the Pearson correlation between two raters' scores can be written as
\begin{equation}
    r_H = \frac{\sigma_T^2}
            {\sqrt{\bigl(\sigma_T^2+\mathrm{Var}(\varepsilon_{\cdot 1})\bigr)
                   \bigl(\sigma_T^2+\mathrm{Var}(\varepsilon_{\cdot 2})\bigr)}}.
    \label{eq:rH_def}
\end{equation}
In contrast, the reliability of the two-rater average is
\begin{equation}
    \rho_Y = \frac{\sigma_T^2}{\sigma_Y^2}
    = \frac{\sigma_T^2}
           {\sigma_T^2 + {(\mathrm{Var}(\varepsilon_{\cdot 1})+\mathrm{Var}(\varepsilon_{\cdot 2}))}/{4}}.
    \label{eq:rhoY_two_raters_general}
\end{equation}
Therefore, it suffices to confirm that
\begin{equation}
    \rho_Y - r_H^2 \ge 0 
    \Longleftrightarrow \frac{3}{4}\sigma_T^2(\mathrm{Var}(\varepsilon_{\cdot 1})+\mathrm{Var}(\varepsilon_{\cdot 2}))+\mathrm{Var}(\varepsilon_{\cdot 1})\mathrm{Var}(\varepsilon_{\cdot 2}) \ge 0,
\end{equation}
which holds because all variance terms are nonnegative.

Finally, we prove the relation $\kappa_H \lesssim \kappa_{\mathrm{HL}}\le \kappa_{\max}$ introduced in Eq.~\eqref{eq:all_index}.
First, because $\rho_1\le 1$, we have $\sqrt{\rho_1 \rho_Y} \le \sqrt{\rho_Y}$, which implies $\kappa_{\mathrm{HL}}\le \kappa_{\max}$.
Next, to prove $\kappa_H \lesssim \kappa_{\mathrm{HL}}$, assume homoscedastic rater errors,
$\mathrm{Var}(\varepsilon_{\cdot 1}) = \mathrm{Var}(\varepsilon_{\cdot 2}) = \sigma_{\varepsilon}^2$.
Then Eq.~\eqref{eq:rH_def} yields
$r_H = {\sigma_T^2}/{(\sigma_T^2+\sigma_{\varepsilon}^2)} = \rho_1$,
and Eq.~\eqref{eq:kappaH_le_rH} implies $\kappa_H \approx r_H\,F_H \le r_H$.
Therefore, we have 
\begin{equation}
\kappa_H \approx r_H\,F_H \le r_H = \rho_1.
\end{equation}
Furthermore, Eqs.~\eqref{eq:reliability_def_Y} and \eqref{eq:reliability_def_1} imply $\rho_1 \le \rho_Y$ because $\sigma^2_{\bar{\varepsilon}} = \mathrm{Var}\left(\frac{1}{J}\sum_{j=1}^{J} \varepsilon_{ij}\right) \le \sigma^2_{{\varepsilon}}$.
Thus, $\rho_1 \le \sqrt{\rho_1 \rho_Y}$ holds, and we obtain
\begin{equation}
\kappa_H \le \rho_1 \le \sqrt{\rho_1 \rho_Y} \approx \kappa_{\mathrm{HL}}.
\end{equation}

\section{Details of the Real-Data Application Setup}

In our ASAP experiments, we constructed the target variable as either the mean or the sum of the two raters' scores.
However, the ASAP dataset also provides an original composite score whose construction differs across prompts.
Specifically, for Prompts~1 and~7 the composite score is the sum of the two raters' scores, whereas for Prompts~2--6 it is defined as either the minimum or the maximum of the two raters' scores.
Prompt~8 is also based on the sum of two raters' scores, but uses a special summation rule.
Because our framework assumes that the target variable is the mean or the sum of two raters' scores, the original composite score cannot be used as-is for prompts other than 1 and 7.
Therefore, to align the dataset with our assumptions while preserving the original scoring scale as much as possible, we used the mean of the two raters' scores for Prompts~2--6 and the simple sum for Prompt~8.

Moreover, regarding the rater assignment design in ASAP, the dataset documentation states that two trained raters were assigned, but it does not specify whether they were randomly drawn from a rater pool or formed a fixed rater pair.
Given the size of the dataset, it is unlikely that the same fixed two raters scored all essays.
Accordingly, we estimated reliability under the assumption that two raters were randomly assigned from a rater pool, as described in Section~\ref{sec:qwk_ub}.
However, for robustness we also computed reliability under the alternative assumption of a fixed rater pair using an appropriate reliability estimator.
The resulting differences were negligible (less than $10^{-4}$ for all prompts), indicating that the choice of reliability estimator has little impact on this dataset.

Regarding the ELLIPSE dataset, it is publicly described as having been constructed by randomly assigning two raters from a pool of 26 trained raters, which straightforwardly matches our assumptions.

Finally, as mentioned in Section~\ref{sec:real_data}, the reported correlation and QWK values for NPCR and MTAA are not strictly comparable to our ceiling estimates or our BERT baseline.
For NPCR, the target-score construction differs from ours: NPCR appears to use the dataset-provided composite scores, so its targets for Prompts~2--6 and~8 differ slightly from the reconstructed targets used in this study.
For MTAA, the experimental setup differs: MTAA merges the original training and evaluation splits and then re-splits the data, instead of using the official split of ELLIPSE.

\begin{credits}
\subsubsection{\ackname}
This work was supported by JSPS KAKENHI Grant Numbers 25K00833, 24H00739 and 23K17585.

\subsubsection{\discintname}
The authors have no competing interests to declare that are relevant to the content of this article.

\end{credits}

\bibliographystyle{splncs04}
\bibliography{references}

@book{AesBook2016,
	author = {Mark D. Shermis and Jill C. Burstein},
	publisher = {Routledge},
	title = {Automated Essay Scoring: A Cross-disciplinary Perspective},
	year = {2003}}

@inproceedings{ijcai2019-AesSurvey,
	author = {Ke, Zixuan and Ng, Vincent},
	booktitle = {Proceedings of the International Joint Conference on Artificial Intelligence},
	pages = {6300--6308},
	title = {Automated Essay Scoring: A Survey of the State of the Art},
	year = {2019}}

@article{uto-review2021,
	author = {Uto, Masaki},
	journal = {Behaviormetrika},
	number = {2},
	pages = {459--484},
	title = {A review of deep-neural automated essay scoring models},
	volume = {48},
	year = {2021}}

@article{uto2022learning,
	author = {Uto, Masaki and Okano, Masashi},
	journal = {IEEE Transactions on Learning Technologies},
	number = {6},
	pages = {763-776},
	title = {Learning automated essay scoring models using item-response-theory-based scores to decrease effects of rater biases},
	volume = {14},
	year = {2021}}

@inproceedings{Amorim2018AutomatedES,
	author = {Evelin Amorim and M{\'a}rcia Can{\c c}ado and Adriano Veloso},
	booktitle = {Proceedings of the 2018 Conference of the North {A}merican Chapter of the Association for Computational Linguistics},
	pages = {229--237},
	title = {Automated Essay Scoring in the Presence of Biased Ratings},
	year = {2018}}

@article{Wind2018,
	author = {Stefanie A. Wind and Edward W. Wolfe and George Engelhard Jr. and Peter Foltz and Mark Rosenstein},
	journal = {International Journal of Testing},
	number = {1},
	pages = {27-49},
	title = {The Influence of Rater Effects in Training Sets on the Psychometric Quality of Automated Scoring for Writing Assessments},
	volume = {18},
	year = {2018}}

@inproceedings{chen-he-2013-automated,
	author = {Chen, Hongbo and He, Ben},
	booktitle = {Proceedings of the 2013 Conference on Empirical Methods in Natural Language Processing},
	pages = {1741--1752},
	title = {Automated Essay Scoring by Maximizing Human-Machine Agreement},
	year = {2013}}

@inproceedings{cozma-etal-2018-automated,
	author = {Cozma, M{\u{a}}d{\u{a}}lina and Butnaru, Andrei and Ionescu, Radu Tudor},
	booktitle = {Proceedings of the 56th Annual Meeting of the Association for Computational Linguistics},
	pages = {503--509},
	title = {Automated essay scoring with string kernels and word embeddings},
	year = {2018}}

@inproceedings{taghipour-ng-2016-neural,
	author = {Taghipour, Kaveh and Ng, Hwee Tou},
	booktitle = {Proceedings of the 2016 Conference on Empirical Methods in Natural Language Processing},
	pages = {1882--1891},
	title = {A Neural Approach to Automated Essay Scoring},
	year = {2016}}

@article{Burrows2015,
	author = {Simon Burrows and Iryna Gurevych and Benno Stein},
	journal = {Journal of Artificial Intelligence in Education},
	pages = {60--117},
	title = {The eras and trends of automatic short answer grading},
	volume = {25},
	year = {2015}}

@article{elmassry2025systematic,
	author = {ElMassry, Ahmed M and Zaki, Nazar and AlSheikh, Negmeldin and Mediani, Mohammed},
	journal = {IEEE Access},
	title = {A Systematic Review of Pretrained Models in Automated Essay Scoring},
	year = {2025}}

@article{ramesh2022automated,
	author = {Ramesh, Dadi and Sanampudi, Suresh Kumar},
	journal = {Artificial Intelligence Review},
	number = {3},
	pages = {2495--2527},
	title = {An automated essay scoring systems: a systematic literature review},
	volume = {55},
	year = {2022}}

@article{misgna2024survey,
	author = {Misgna, Haile and On, Byung-Won and Lee, Ingyu and Choi, Gyu Sang},
	journal = {Artificial Intelligence Review},
	number = {2},
	pages = {36},
	title = {A survey on deep learning-based automated essay scoring and feedback generation},
	volume = {58},
	year = {2024}}

@inproceedings{yancey2023rating,
	author = {Kevin P. Yancey and Geoffrey Laflair and Anthony Verardi and Jill Burstein},
	booktitle = {Proceedings of the 18th Workshop on Innovative Use of NLP for Building Educational Applications},
	pages = {576--584},
	title = {Rating Short {L2} Essays on the {CEFR} Scale with {GPT}-4},
	year = {2023}}

@inproceedings{lee2024unleashing,
	author = {Sanwoo Lee and Yida Cai and Desong Meng and Ziyang Wang and Yunfang Wu},
	booktitle = {Findings of the Association for Computational Linguistics},
	pages = {181--198},
	title = {Unleashing Large Language Models' Proficiency in Zero-shot Essay Scoring},
	year = {2024}}

@inproceedings{mansour2024can,
	author = {Watheq Ahmad Mansour and Salam Albatarni and Sohaila Eltanbouly and Tamer Elsayed},
	booktitle = {Proceedings of the 2024 Joint International Conference on Computational Linguistics, Language Resources and Evaluation},
	pages = {2777--2786},
	title = {Can Large Language Models Automatically Score Proficiency of Written Essays?},
	year = {2024}}

@book{Bishop:2006,
	address = {New York},
	author = {Christopher M. Bishop},
	publisher = {Springer},
	title = {Pattern Recognition and Machine Learning},
	year = {2006}}

@article{mcgraw1996forming,
	author = {McGraw, Kenneth O and Wong, Seok P},
	journal = {Psychological Methods},
	number = {1},
	pages = {30--46},
	title = {Forming inferences about some intraclass correlation coefficients},
	volume = {1},
	year = {1996}}

@article{barnhart2007overall,
	author = {Barnhart, Huiman X and Haber, Michael J and Lin, Lawrence I},
	journal = {Biometrics},
	number = {4},
	pages = {1099--1106},
	title = {Overall concordance correlation coefficient for evaluating agreement among multiple observers},
	volume = {63},
	year = {2007}}

@inproceedings{npcr2022,
	author = {Xie, Jiayi and Cai, Kaiwei and Kong, Li and Zhou, Junsheng and Qu, Weiguang},
	booktitle = {Proceedings of the 29th International Conference on Computational Linguistics},
	pages = {2724--2733},
	title = {Automated Essay Scoring via Pairwise Contrastive Regression},
	year = {2022}}

@inproceedings{ellipse_aied24,
	author = {Chen, Shigeng and Lan, Yunshi and Yuan, Zheng},
	booktitle = {Proceedings of the International Conference on Artificial Intelligence in Education},
	pages = {276--283},
	title = {A Multi-task Automated Assessment System for Essay Scoring},
	year = {2024}}

@article{EnrightQuinlan2010,
  author  = {Enright, Mary K. and Quinlan, Thomas},
  title   = {Complementing Human Judgment of Essays Written by English Language Learners with e-rater Scoring},
  journal = {Language Testing},
  volume  = {27},
  number  = {3},
  pages   = {317--334},
  year    = {2010}
}

@article{RamineniWilliamson2013,
  author  = {Ramineni, Chaitanya and Williamson, David M.},
  title   = {Automated essay scoring: Psychometric guidelines and practices},
  journal = {Assessing Writing},
  volume  = {18},
  number  = {1},
  pages   = {25--39},
  year    = {2013}
}

@article{shibata2025cross,
  title={Cross-Prompt Automated Essay Scoring via Reinforcement Learning-Based Data Valuation},
  author={Shibata, Takumi and Uto, Masaki},
  journal={IEEE Access},
  volume={13},
  pages={184792--184808},
  year={2025}}

@inproceedings{yamaura2023neural,
  title={Neural automated essay scoring considering logical structure},
  author={Yamaura, Misato and Fukuda, Itsuki and Uto, Masaki},
  booktitle={Proceedings of the International Conference on Artificial Intelligence in Education},
  pages={267--278},
  year={2023}}

@article{uto2020item,
  title={A generalized many-facet {R}asch model and its {B}ayesian estimation using {H}amiltonian {M}onte {C}arlo},
  author={Uto, Masaki and Ueno, Maomi},
  journal={Behaviormetrika},
  volume={47},
  number={2},
  pages={469--496},
  year={2020},
  publisher={Springer}}

@book{MFRM2016BOOK,
  author = {Eckes, Thomas},
  publisher = {Peter Lang Pub. Inc.},
  title = {Introduction to Many-Facet {Rasch} Measurement: Analyzing and Evaluating Rater-Mediated Assessments},
  year = {2023}}

@book{brennan2001generalizability,
  title={Generalizability Theory},
  author={Brennan, Robert L.},
  year={2001},
  publisher={Springer-Verlag},
  address={New York}}

\end{document}